\documentclass[runningheads]{llncs}
\usepackage{graphicx}
\usepackage{amsmath,amssymb} 
\usepackage{color}
\usepackage{booktabs}
\usepackage{multirow}
\usepackage{array}
\usepackage{rotating}
\usepackage{wrapfig}
\usepackage{caption}
\usepackage{marvosym}
\usepackage{hyperref}
\makeatletter
\newcommand{\printfnsymbol}[1]{%
  \textsuperscript{\@fnsymbol{#1}}%
}
\makeatother
\newcommand{\etal}{\emph{et al.}~}

\begin{document}
\pagestyle{headings}
\mainmatter

\def\ACCV22SubNumber{367}  

\title{Multi-Scale Wavelet Transformer for Face Forgery Detection} 
\titlerunning{MSWT for Face Forgery Detection}

\author{Jie Liu\thanks{Equal contribution.}
\and Jingjing Wang\printfnsymbol{1}
\and Peng Zhang
\\
\and Chunmao Wang
\and Di Xie
\and Shiliang Pu\textsuperscript{\Letter}
}
\authorrunning{Liu, J. et al.}
%
\institute{Hikvision Research Institute\\
\email{\{liujie54, wangjingjing9, zhangpeng45, wangchunmao, xiedi, pushiliang.hri\}@hikvision.com}}
%
\maketitle

\begin{abstract}
	Currently, many face forgery detection methods aggregate spatial and frequency features to enhance the generalization ability and gain promising performance under the cross-dataset scenario. However, these methods only leverage one level frequency information which limits their expressive ability. To overcome these limitations, we propose a multi-scale wavelet transformer framework for face forgery detection. Specifically, to take full advantage of the multi-scale and multi-frequency wavelet representation, we gradually aggregate the multi-scale wavelet representation at different stages of the backbone network. To better fuse the frequency feature with the spatial features, frequency-based spatial attention is designed to guide the spatial feature extractor to concentrate more on forgery traces. Meanwhile, cross-modality attention is proposed to fuse the frequency features with the spatial features. These two attention modules are calculated through a unified transformer block for efficiency. A wide variety of experiments demonstrate that the proposed method is efficient and effective for both within and cross datasets.
	
\end{abstract}

\section{Introduction}
Due to the various image-editing software and publicly available deep generator models, it is easy to manipulate existing faces and make forged faces very realistic and indistinguishable from genuine ones. These photo-realistic fake faces may be abused for malicious purposes, raising severe security and privacy issues in our society. Therefore, it is extremely necessary to develop effective methods for face forgery detection. To defend against the possible malicious usage of face forgery, various face forgery detection methods have been proposed. Previous researchers \cite{ferrara2012image,pan2012exposing} mainly designed methods based on texture artifacts caused by the face forgery techniques in the spatial domain. Due to the fast evolution of face forgery techniques, these artifacts are gradually concealed. Therefore, although these methods achieved high within-dataset detection accuracy, their performance dropped severely in the cross-dataset scenario, especially when confronted with new face forgery methods.

\begin{table}[htpb]
	\centering
	\begin{tabular}{c|c|c|c|c|c}
		\toprule
		Level       & Sub-bands & Deepfakes(DF) & Face2Face(F2F) & FaceSwap(FS) & NeuralTextures(NT) \\
		\hline
		\hline
		-           & Ori-Img       & 1.301 & 1.092 & 1.307 & 1.296 \\
		\hline
		\multirow{4}{*}{Level-1} & LL        & 1.281 & 1.008 & 1.265 & 1.208 \\
		& LH        & 2.688 & 2.709 & 2.970 & 2.959 \\
		& HL        & 2.716 & 2.778 & 2.720 & 2.857 \\
		& HH        & 2.582 & 2.914 & 3.258 & 2.758 \\
		\hline
		\multirow{4}{*}{Level-2} & LL        & 1.208 & 0.958 & 1.165 & 1.162 \\
		& LH        & 2.817 & 2.840 & 2.882 & 3.106 \\
		& HL        & 2.598 & 2.686 & 2.549 & 2.871 \\
		& HH        & 3.184 & 2.929 & 3.162 & 3.127 \\
		\hline
		\multirow{4}{*}{Level-3} & LL        & 1.189 & 1.055 & 1.136 & 1.246 \\
		& LH        & 2.473 & 2.493 & 2.510 & 2.826 \\
		& HL        & 2.135 & 2.409 & 2.166 & 2.837 \\
		& HH        & 2.774 & 2.917 & 2.936 & 2.985 \\
		\bottomrule
	\end{tabular}
	\caption{EMD of multi-level frequency components. Cropping the face in the first frame of every video in FF++ dataset, and then calculating the EMD of the original images or sub-bands frequency features between the fake and corresponding real images. These sub-bands are obtained by three level discrete wavelet transform.}
	\label{table:dwt_emd}
\end{table}

To make the algorithm generalize well to unseen forgery methods, recently, many face forgery detection methods attempt to aggregate information from frequency domains. Yu \etal \cite{yu2020mining} utilized channel difference images and the spectrum obtained by DCT to detect fake faces. Other researchers leveraged Discrete Fourier Transform (DFT) \cite{SPSL} and Discrete Cosine Transform (DCT) and block DCT \cite{F3-Net} for frequency information extracting.
However, these methods only utilized one level frequency information. And we found that multi-level frequency features have more discriminable details between real and fake images. Only using one level frequency may be less effective for extracting the abundant frequency information, which limits the expressive ability of the obtained features.
As we all know, Discrete Wavelet Transform (DWT) is often used to obtain multi-level frequency, so we choose Haar DWT to extract frequency features. The filter $f_{LL}$, $f_{LH}$, $f_{HL}$, and $f_{HH}$ of DWT are $\frac{1}{2}
\begin{bmatrix}
1 & 1\\
1 & 1
\end{bmatrix}
$, $\frac{1}{2}
\begin{bmatrix}
1 & 1\\
-1 & -1
\end{bmatrix}
$, $\frac{1}{2}
\begin{bmatrix}
1 & -1\\
1 & -1
\end{bmatrix}
$, and $\frac{1}{2}
\begin{bmatrix}
1 & -1\\
-1 & 1
\end{bmatrix}
$, and they are used to calculate the frequency (LL, LH, HL, HH) of an image $I$. The LL, LH, HL, and HH are defined as $LL = f_{LL} * I$, $LH = f_{LH} * I$, $HL = f_{HL} * I$, $HH = f_{HH} * I$. DWT divides an image into four frequency components with half resolution of the original image: a low-frequency component (LL) and three high-frequency components (LH, HL, HH).
And the LL can be further decomposed into four frequency components recursively. In this way, we can get multi-level wavelet representations. Earth Mover's Distance (EMD) \cite{rubner2000earth} is used to measure the dissimilarity between two multidimensional distributions, whose formula is defined in \cite{rubner2000earth}. The total EMD distance of FF++ dataset is calculated by three level frequency components between the real and fake data, whose results are shown in Table \ref{table:dwt_emd}. We observe that the distance of high-frequency information between real and fake facial images is bigger than low-frequency one at each level, which demonstrates that different level high frequencies are all useful so that fusing multi-level high frequencies can make the representations more expressive for face forgery detection.

\begin{figure}[htpb]
	\centering
	\begin{tabular}{c@{\hskip 0.5mm}c@{\hskip 1mm}c@{\hskip 1mm}c@{\hskip 1mm}c|@{\hskip 1mm}c@{\hskip 1mm}c@{\hskip 1mm}c@{\hskip 1mm}c}
		\rotatebox{90}{~~~~fake}&
		\includegraphics[width=0.113\textwidth]{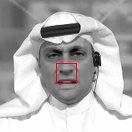}
		&
		\includegraphics[width=0.113\textwidth]{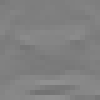}
		&
		\includegraphics[width=0.113\textwidth]{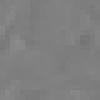}
		&
		\includegraphics[width=0.113\textwidth]{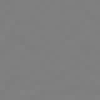}
		
		&
		\includegraphics[width=0.116\textwidth]{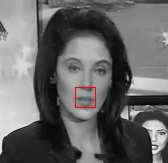}
		&
		\includegraphics[width=0.113\textwidth]{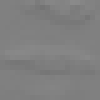}
		&
		\includegraphics[width=0.113\textwidth]{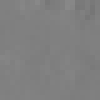}
		&
		\includegraphics[width=0.113\textwidth]{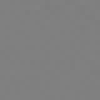}
		
		\\
		\rotatebox{90}{~~~~real}&
		\includegraphics[width=0.113\textwidth]{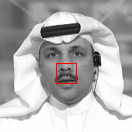}
		&
		\includegraphics[width=0.113\textwidth]{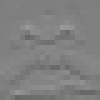}
		&
		\includegraphics[width=0.113\textwidth]{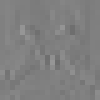}
		&
		\includegraphics[width=0.113\textwidth]{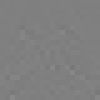}
		
		&
		\includegraphics[width=0.116\textwidth]{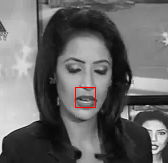}
		&
		\includegraphics[width=0.113\textwidth]{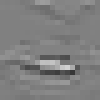}
		&
		\includegraphics[width=0.113\textwidth]{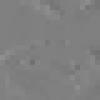}
		&
		\includegraphics[width=0.113\textwidth]{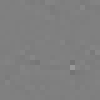}
		\\
		
		\rotatebox{90}{~~~~fake}&
		\includegraphics[width=0.113\textwidth]{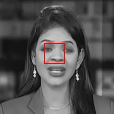}
		&
		\includegraphics[width=0.113\textwidth]{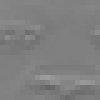}
		&
		\includegraphics[width=0.113\textwidth]{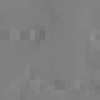}
		&
		\includegraphics[width=0.113\textwidth]{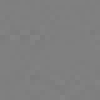}
		
		&
		\includegraphics[width=0.113\textwidth]{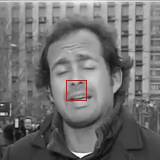}
		&
		\includegraphics[width=0.113\textwidth]{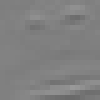}
		&
		\includegraphics[width=0.113\textwidth]{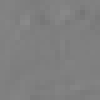}
		&
		\includegraphics[width=0.113\textwidth]{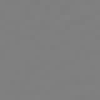}
		\\
		
		\rotatebox{90}{~~~~real}&
		\includegraphics[width=0.113\textwidth]{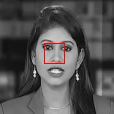}
		&
		\includegraphics[width=0.113\textwidth]{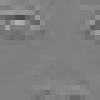}
		&
		\includegraphics[width=0.113\textwidth]{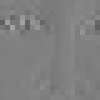}
		&
		\includegraphics[width=0.113\textwidth]{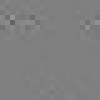}
		
		&
		\includegraphics[width=0.113\textwidth]{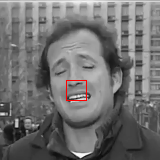}
		&
		\includegraphics[width=0.113\textwidth]{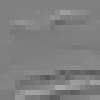}
		&
		\includegraphics[width=0.113\textwidth]{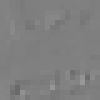}
		&
		\includegraphics[width=0.113\textwidth]{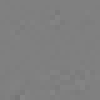}
		\\
		~ & Gray & LH & HL & HH & Gray & LH & HL & HH
		\\
	\end{tabular}
	\caption{High-frequency sub-bands are obtained by DWT. The images in the 1st and 3rd lines are fake images, and the others are real images. In this figure, we show the fake facial images and their corresponding real images. The 1st and 5th column are the gray images, and the column 2 to 4 and 6 to 8 are high frequency sub-bands corresponding to the cropped red box. The forged pixels have fewer high-frequency details (LH, HL, HH) compared with the real ones.}
	\label{fig:dwt_ex}
\end{figure}

\begin{figure}[htpb]
	\centering
	\begin{tabular}{c@{\hskip 1mm}c@{\hskip 1mm}c@{\hskip 1mm}c@{\hskip 1mm}c@{\hskip 1mm}c@{\hskip 1mm}c@{\hskip 1mm}c@{\hskip 1mm}c@{\hskip 1mm}c}
		\includegraphics[width=0.09\textwidth]{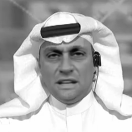}
		&
		\includegraphics[width=0.09\textwidth]{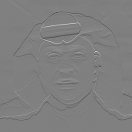}
		&
		\includegraphics[width=0.09\textwidth]{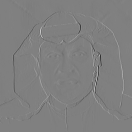}
		&
		\includegraphics[width=0.09\textwidth]{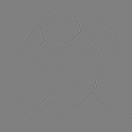}
		&
		\includegraphics[width=0.09\textwidth]{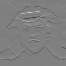}
		&
		\includegraphics[width=0.09\textwidth]{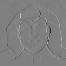}
		&
		\includegraphics[width=0.09\textwidth]{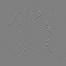}
		&
		\includegraphics[width=0.09\textwidth]{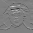}
		&
		\includegraphics[width=0.09\textwidth]{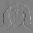}
		&
		\includegraphics[width=0.09\textwidth]{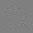}
		\\
		\includegraphics[width=0.09\textwidth]{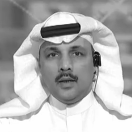}
		&
		\includegraphics[width=0.09\textwidth]{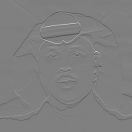}
		&
		\includegraphics[width=0.09\textwidth]{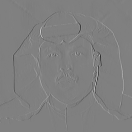}
		&
		\includegraphics[width=0.09\textwidth]{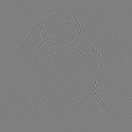}
		&
		\includegraphics[width=0.09\textwidth]{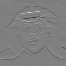}
		&
		\includegraphics[width=0.09\textwidth]{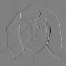}
		&
		\includegraphics[width=0.09\textwidth]{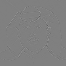}
		&
		\includegraphics[width=0.09\textwidth]{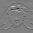}
		&
		\includegraphics[width=0.09\textwidth]{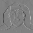}
		&
		\includegraphics[width=0.09\textwidth]{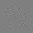}
		\\
		Gray & LH\_L1 & HL\_L1 & HH\_L1 & LH\_L2 & HL\_L2 & HH\_L2 & LH\_L3 & HL\_L3 & HH\_L3
		\\
	\end{tabular}
	\caption{The images in 1st and 2nd lines are the fake and the real facial images, respectively. Columns 2 to 10 are level 1, 2, and 3 high frequency (LH, HL, and HH) sub-bands by DWT. There is more details in lower levels, and more global semantic structure in higher levels.}
	
	\label{fig:dwt_three_level}
\end{figure}

We also visualize the examples of the real and fake high frequency by DWT in Figure \ref{fig:dwt_ex} and \ref{fig:dwt_three_level}. In Figure \ref{fig:dwt_ex}, we enlarge the local region of the first level DWT, so we can see that there are more details in low-level high frequency. Figure \ref{fig:dwt_three_level} shows the whole high-frequency sub-bands of the three-level DWT, and there is more global semantic information in high-level frequency. So the low-level and high-level high-frequency features are all important for facial forgery detection.

Taking the above considerations, we take the multi-scale analysis of wavelet decomposition into consideration and propose a multi-scale wavelet transformer framework for face forgery detection named MSWT. Specifically, we gradually aggregate the multi-scale wavelet features at different stages of the backbone network to take full advantage of multi-level high-frequency representation. To better fuse the frequency feature with the spatial features, frequency-based spatial attention is designed to guide the spatial feature extractor to concentrate more on forgery traces. Meanwhile, cross-modality attention is proposed to fuse the RGB spatial features and the frequency features. These two attention modules are calculated through a unified transformer block for efficiency named frequency and spatial feature fusion (FSF) module. The main contributions are summarized as follows:
\begin{itemize}
	\item To make full use of frequency features, we are the first to utilize the multi-scale properties of wavelet decomposition to improve the feature fusion of spatial and frequency domains, and propose a multi-scale wavelet transformer framework for face forgery detection.
	\item To better capture the manipulation trace, frequency-based spatial attention is designed to guide spatial feature extractor to focus on forgery regions.
	\item To better fuse the frequency features with the RGB spatial features, cross-modality attention is introduced.
	\item Experiments demonstrate that the proposed method works well on both within-dataset and cross-dataset testing compared with other approaches.
\end{itemize}

\section{Related Work}
\subsection{Forgery Detection}
\textbf{Forgery Detection based on Spatial Feature.}
In order to resist manipulated faces and protect media security, many forgery detection algorithms have been proposed in academia. Because deep learning can learn good feature representation, some methods are proposed to extract RGB spatial features based on deep learning. These approaches mainly include consisrency-based \cite{zhao2020learning}, attention-based \cite{MADFD}, and domain generalization methods \cite{sun2021domain}. Zhao \etal \cite{MADFD} proposed a method based on multi-attention and textural feature enhancement to enlarge artifacts in shallow features and capture discriminative details for face forgery detection, fusing the low-level and high-level features by attention maps. Zhao \etal \cite{zhao2020learning} proposed patch-wise consistency learning between patches from the feature maps, which utilizes consistency loss to learn and optimize the consistency of the patches from real or fake regions. Wodajo \etal \cite{wodajo2021deepfake} proposed a convolutional vision transformer for deepfake video detection, and the network consists of a convolutional neural network (CNN) and a vision transformer (ViT). The ViT processes the features learned by CNN and then exports the classification results. Different from Wodajo \etal, we only utilize transformer encoder as the attention module to fuse high-frequency components and RGB spatial features.

\textbf{Forgery Detection based on Frequency.}
Because of the effectiveness of frequency information for forgery detection, some proposed networks combined the spatial features with frequency information. Li \etal \cite{FDF} offered a single-center loss to learn the frequency-aware features and used the Discrete Cosine Transform (DCT) transform to get the frequency representation. Qian \etal \cite{F3-Net} proposed F3-Net for face forgery detection, utilizing DCT and block DCT to calculate the global and block frequency information. Zhu \etal \cite{Zhu_2021_CVPR} also utilized frequency and spatial feature for forgery detection, with a two-stream architecture for RGB image and high-frequency component respectively. SPSL \cite{SPSL} was proposed by Liu \etal, in which DFT is applied to extract the phase spectrum as the high-frequency representation. However, these methods only use one level frequency information, discarding the valuable information in multi-level frequency information.

\subsection{Wavelet in Computer Vision Tasks}

For images, DWT obtains frequency and spatial components simultaneously, as shown in Figure \ref{fig:dwt_ex}. Besides, wavelet transform has the characteristics of multi-resolution analysis, by which multi-scale frequency feature representation calculated by DWT has tremendous significance in computer vision tasks. For example, DWT is used in image depth prediction \cite{SID_wavelet}, image denoising \cite{malfait1997wavelet,pizurica2002joint},  restoration \cite{figueiredo2003algorithm}, compression \cite{meyer2000fast}, and fusion \cite{shi2005wavelet,pajares2004wavelet}, achieving good performance at that time.

\section{Proposed Method}

\subsection{Overview of the Structure}
\begin{figure}[htbp]
	\centering
	\includegraphics[width=1.0\columnwidth]{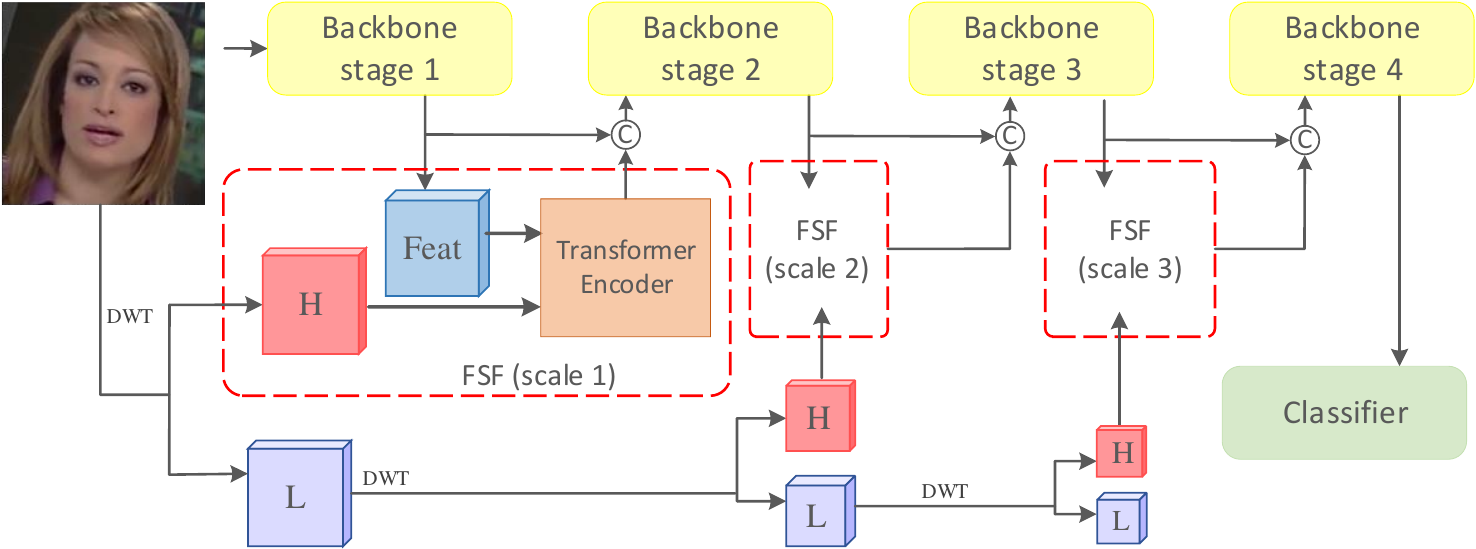}
	\caption{The architecture of the proposed method. The backbone is Xception \cite{chollet2017xception}. This backbone is split into four stages according to its features' resolution. The backbone extracts the features from the original image, and DWT is used to divide the input image into the low-frequency ($L$) and high-frequency ($H$) components at each channel (RGB). The frequency and spatial feature fusion (FSF) module as shown in the red dashed rectangle takes high-frequency information $H$ and the spatial features $Feat$ extracted from the corresponding backbone stage as inputs.}
	\label{fig:architecture}
\end{figure}

For face forgery detection, we propose a multi-scale wavelet transformer architecture. The comprehensive framework is depicted in Figure~\ref{fig:architecture}, which takes the RGB image and the multi-level high-frequency representations via DWT as inputs and fuses high-frequency features and the RGB spatial features with the proposed fusion module. According to the different input sources, the network is divided into the RGB branch which takes the RGB image as input, and the high frequency branches which take the wavelet-based high frequency representations as input. For the RGB branch, following most works, we take Xception \cite{chollet2017xception} as the feature extractor. According to the resolution of feature maps, we split it into four convolutional feature extracting stages and one classifier, shown as the yellow and green blocks in Figure \ref{fig:architecture}.

For the high frequency branches, we take Haar wavelet \cite{Haar1910} to get the multi-level high-frequency representations (shown as the red blocks in Figure \ref{fig:architecture}) as input due to its simplicity and efficiency. DWT can divide a gray image into four frequency components which consist of one low-frequency (LL) and three high-frequency components (LH, HL, and HH) with half resolution of the original image. We do DWT for each channel of the RGB image respectively. Therefore, each frequency component consists of three channel maps corresponding to the red, green, and blue channel of the RGB image.
As shown in the high frequency branch of Figure \ref{fig:architecture}, taking the three high-frequency components ``H'' as the frequency representations at the current level and using the low-frequency component ``L" to do further wavelet decompositions, we can get the multi-level frequency representations. As analyzed in the introduction, frequency representations at different levels contain different useful information. In our method, we use high-frequency representations of the first three levels.

To fuse the information among different branches, we aggregate the information from the high frequency branches into the RGB branch using the proposed light-weight  Frequency and Spatial Feature Fusion (FSF) module at three levels for efficiency, shown as the red dashed rectangle of Figure \ref{fig:architecture}. The information of the three high frequency branches is fused into the corresponding stages of the RGB branch to match their resolutions. Except for matching resolution, another reason is that, as analyzed in the introduction the low level frequency representations contain more details, while the high level frequency representations contain more global structure information, which has similar meanings to the features in the multiple stages of the RGB branch. The FSF module is based on the transformer and its details are described in the next subsection \ref{sec:FSF}. Finally, the enhanced features are input into the classifier, and the whole network is optimized end-to-end using the cross-entropy loss.

\subsection{Frequency and Spatial Feature Fusion Module}\label{sec:FSF}

The structure of frequency and spatial feature fusion (FSF) module is shown in Figure \ref{fig:FFM} (a). It takes features $Feat$ from the RGB branch and high-frequency representations $H$ at the corresponding level as inputs to fuse the high frequency and RGB spatial features. FSF module consists of two attention blocks. One block is used to enhance spatial feature maps $Feat$ with the high-frequency guided attention which is denoted as frequency-based spatial attention (FSA). The other block is used to fuse the high-frequency information into $Feat$ with cross-modality attention (CMA). Finally, the outputs of these two blocks are concatenated as the output of FSF module as shown in Figure \ref{fig:FFM} (a).
The high-frequency representations $H$ ($LH, LH, HH$) are obtained by DWT via the original image, while $Feat$ are RGB spatial features obtained from the RGB branch. To match the meaning of $Feat$, each part of $H$ is first processed with the corresponding convolutional block. Then they are concatenated and fused with another convolution to get the output high-frequency feature maps $F_H$. This operation is defined as:
\begin{equation}\label{eq:wconv}
	F_H = f_{combine} \left( Concat \left(  f_{conv}^{LH} \left( LH \right) , f_{conv}^{HL}  \left( HL \right) , f_{conv}^{HH} \left( HH \right) \right) \right)
\end{equation}
where $f_{conv}$ is convolution block used to process the high-frequency features, and then the processed high-frequency features are concatenated with $Concat$. $f_{combine}$ fuses all high-frequency features with another convolution operation. The total operation process is shown in the blue dashed rectangle of Figure \ref{fig:FFM} (a).

On the other hand, to match the channel dimension of feature maps $F_H$, a down-channel convolution is performed on the RGB spatial features $Feat$. The down-channel feature is defined as:
\begin{equation}\label{eq:downc}
	F_S = f_{DownConv} \left( Feat \right)
\end{equation}
where $f_{DownConv}$ means the down-channel convolution operation.

The high frequency features $F_H$ and spatial features $F_S$ are fused using frequency-based spatial attention (FSA) and cross-modality attention (CMA), which are the yellow and the green rectangular boxes in Figure \ref{fig:FFM} (a). The first attention is used to guide the RGB spatial feature learning with high frequency information, and the second one is used to fuse information from high frequency features and spatial features. They are illustrated in the following subsection in detail.

\subsubsection{Frequency-based Spatial Attention}
\begin{figure}[htbp]
	\centering
	\centering
	\begin{tabular}{c@{\hskip 1mm}c}
		\includegraphics[width=0.385\columnwidth]{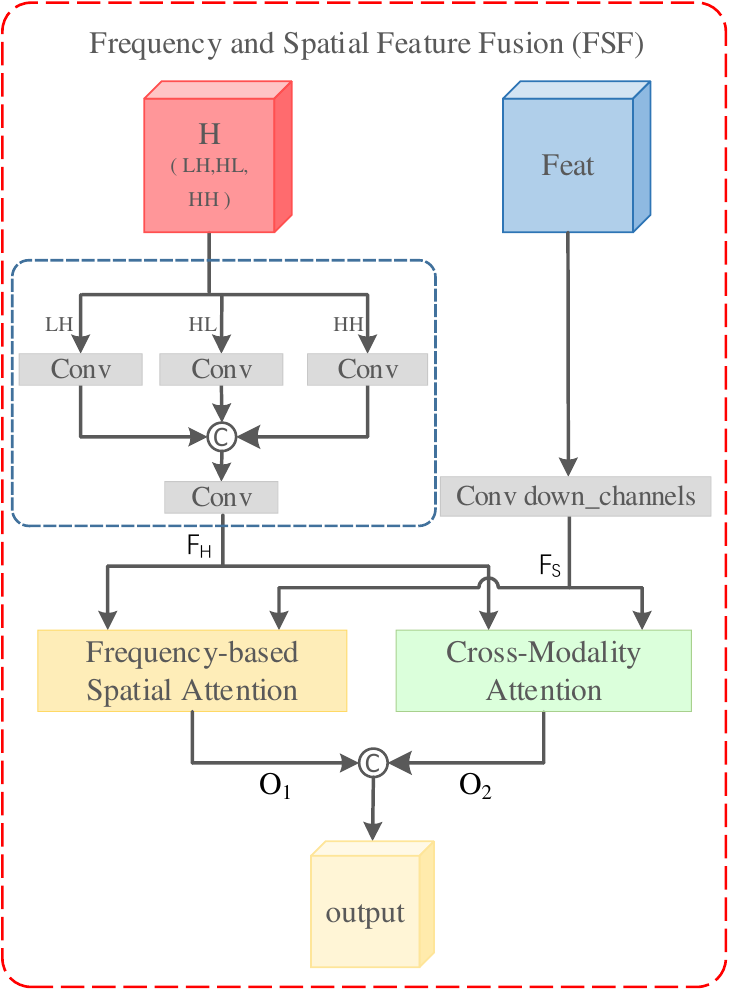}
		&
		\includegraphics[width=0.615\columnwidth]{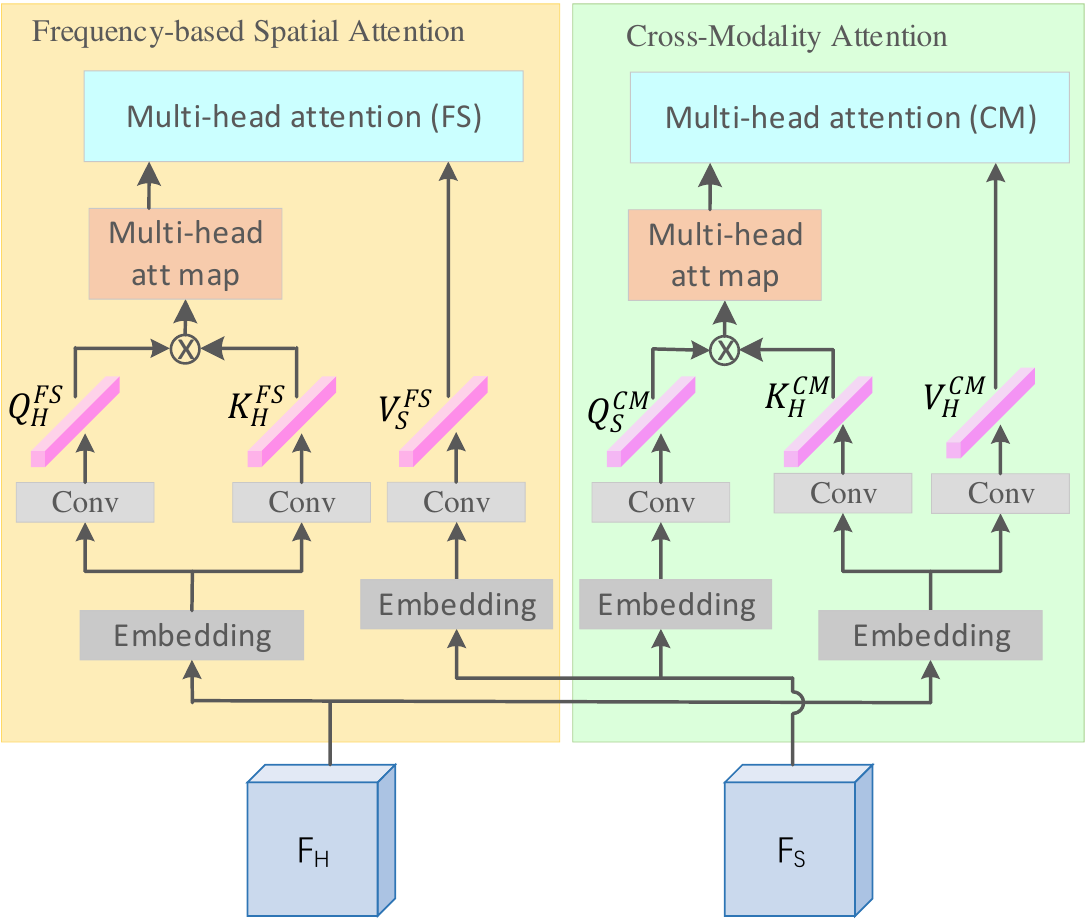}
		\\
		(a) & (b) \\
	\end{tabular}
	\caption{(a) FSF module. (b) The architectures of frequency-based spatial attention (FSA) and cross-modality attention (CMA). The left block is FSA, in which Q and K are calculated with high-frequency features and V is obtained by spatial features. The right one is CMA, in which Q is calculated with spatial features, and the others are obtained with high-frequency features.}
	\label{fig:FFM}
\end{figure}

The area of the manipulated region usually contains the total face or an expression region. For example, the main region of the original face is replaced by another face when using Deepfake as the forgery method. Therefore, leveraging the long-term relationship is helpful to enhance the representation ability of the features. We take transformer to model this relation. However, using the spatial features $F_S$ to calculate the self-attention may not be the best choice, since the self-attention based on $F_S$ is more likely disturbed by the appearance details. Therefore, we propose to use $F_H$ to calculate the attention, since the manipulated regions have similar high-frequency forgery traces.
The architecture of frequency-based spatial attention is shown in the left rectangular box in the Figure. \ref{fig:FFM} (b).
First, the attention query $Q_H^{FS}$ and key $K_H^{FS}$ are calculated by embedding the high-frequency features $F_H$. And the value $V_S^{FS}$ is obtained by embedding spatial features $F_S$. The output is named $O_{1}$, which is defined as:
\begin{equation}\label{eq:fsa}
	O_1 = MHA \left( Q_H^{FS}, K_H^{FS}, V_S^{FS} \right)
\end{equation}
where $MHA$ is the multi-head attention of the vision transformer, and the superscript $FS$ means the vector of Frequency-based Spatial Attention.

\subsubsection{Cross-Modality Attention}
The cross-modality attention is used to fuse the information from the high-frequency representations and RGB spatial features.
The architecture of cross-modality attention is shown in the right rectangular box in Figure \ref{fig:FFM} (b). The key $K_H^{CM}$ and value $V_H^{CM}$ are calculated by embedding the high-frequency features $F_H$, and the query $Q_S^{CM}$ is obtained by embedding spatial features $F_S$. And the output of cross-modality attention is $O_{2}$, which is defined as follows:
\begin{equation}\label{eq:cma}
	O_2 = MHA \left( Q_S^{CM}, K_H^{CM}, V_H^{CM} \right)
\end{equation}
where the superscript $CM$ means the vector is in Cross-Modality Attention.

By the above two attentions, the features are enhanced through the frequency information from two different aspects. On the one hand, the features are enhanced by fusing the information of the relative regions in the spatial domain with high frequency guiding. On the other hand, the features are enhanced by fusing the information of the relative regions at the high frequency domain with cross-modality attention. Finally, we concatenate $O_{1}$ and $O_{2}$ as the output of FSF module, and then send them and the original RGB spatial features of this stage of RGB branch into the next RGB branch stage as shown in Figure \ref{fig:architecture}.

\section{Experiments and Analysis}

\subsection{Datasets}

There are four popular datasets used in experiments, FaceForensics++ (FF++) \cite{FF++}, Celeb-DF \cite{celebdf}, FFIW \cite{zhou2021face}, and WildDeepfake (WDF) \cite{zi2020wilddeepfake} datasets.
\textbf{FF++: } \cite{FF++}
It consists of four manipulation methods, which are DeepFakes \cite{DeepFakes}, Face2Face \cite{thies2016face2face}, FaceSwap \cite{FaceSwap}, and NeuralTexture \cite{thies2019deferred}. And the videos have three compression settings: raw, high quality (c23), and low quality (c40). FF++ is widely used in the forgery detection task, including 1000 videos for each manipulation method and real data. According to the official split, we extract frames from 720, 140, and 140 videos as training, validation, and testing datasets respectively. We get frames from videos by FFmpeg. We use all frames and 300 frames (each video) in the training and testing phases.
\textbf{Celeb-DF: } \cite{celebdf}.
Li \etal proposed Celeb-DF dataset for forgery detection. Before long, they added more videos into Celeb-DF, and then the Celeb-DF V2 dataset appeared. Celeb-DF V2 is the most popular cross-dataset in forgery detection. Therefore, we also use Celeb-DF V2 as one of the cross-dataset evaluations. For each test video, we extract one frame for every five frames in each video.
To evaluate methods in more realistic scenarios, two new public large-scale deepfake datasets in the wild \textbf{FFIW} \cite{zhou2021face} and \textbf{WDF} \cite{zi2020wilddeepfake} are used for within- and cross-domain evaluations. FFIW is a large scale deepfake dataset with high quality, and the fake videos are photo-realistic and close to the real world. WDF collects videos purely from the internet which is more diverse and closer to the real-world deepfakes.

\begin{table*}[htbp]
	\centering
	\begin{tabular}{c|cc|cc|cc|cc|cc|cc}
		\toprule
		\multirow{2}{*}{Methods} & \multicolumn{2}{c|}{DF\_c40} & \multicolumn{2}{c|}{F2F\_c40} & \multicolumn{2}{c|}{FS\_c40} & \multicolumn{2}{c|}{NT\_c40} & \multicolumn{2}{c|}{FF++\_c23}   & \multicolumn{2}{c}{FF++\_c40} \\
		\cline{2-13}
		& ACC & AUC & ACC & AUC & ACC & AUC & ACC & AUC & ACC & AUC & ACC & AUC\\
		\hline
		\hline
		Two-branch \cite{masi2020two}
		& - & - & - & -
		& - & - & - & -
		& 96.43 & - & 86.34 & - \\
		
		MADFD~\cite{MADFD}
		& - & - & - & -
		& - & - & - & -
		&  -     & 98.9  & -     & 87.2 \\
		
		Xception~\cite{chollet2017xception}
		& 95.1 & 99.0& 83.4 & 93.7
		& 92.0 & 97.4& 77.8 & 84.2
		& 95.7 & 96.3& 86.8 & 89.3\\
		
		F3Net~\cite{F3-Net}
		& 97.9 & - & 95.3 & -
		& 96.5  & -  & 83.3 & -
		&  97.5 & 98.1  & 90.4 & 93.3 \\
		
		SPSL~\cite{SPSL}
		& 93.4 & 98.5&86.0 & 94.6
		& 92.2 & 98.1  & 76.7 & 80.4
		&  -    & 95.3  & 80.3 & 82.8 \\
		
		FADFL~\cite{li2021frequency}
		& - & - & - & -
		& - & - & - & -
		& 96.6 & 99.3 & 89.0 & 92.4 \\
		
		GFFD~\cite{Luo_2021_CVPR}
		& \textbf{98.6} & - & \textbf{95.7} & -
		& 92.9  & -     & - & -
		& - & - & - & -\\
		ours
		& 97.8 & 99.7 & 94.6 & 98.5
		& \textbf{98.2}  & \textbf{99.1} & \textbf{86.6} & \textbf{93.8}
		& \textbf{98.6} & \textbf{99.8} &  \textbf{94.9} & \textbf{98.6} \\
		\bottomrule
	\end{tabular}
	\caption{The results of within-dataset evaluation on FF++ dataset (video level). In this table, we show the results of each manipulation method (DF, FF, FS, and NT) based on c40. The last two columns show the results of c23 and c40 of FF++ dataset. The metrics are ACC and AUC.}
	\label{table:self_split}
\end{table*}
\subsection{Experiment Details}
We crop the face and resize them to $384 \times 384$ according to key points by MTCNN \cite{zhang2016joint} as the input of the network. The model of RGB branch is initialized with the parameters pre-trained on ImageNet. In the training phase, we only use random horizontal flip as data augmentation, because we don't want other data augmentations to interfere with the final experimental results. The loss function of the model is cross-entropy loss. The batch size is set to $24$. For training, we adopt AdamW \cite{adamw} optimizer to optimize the total network, whose coefficients are set to $0.9$ and $0.999$ as default. The learning rate is initialized as $0.0001$ and decreases by $0.5$ with StepLR schedule for each $6 \times 10^{4}$ iterations, and the total number of iteration is $1.5 \times 10^5$. For attention layers, the number of heads is set to 1, 2, 5 empirically, and the embedding dimension is set to be 64, 128, 320 at attention level 1 to 3 respectively.

\subsection{Results and Analysis}
\textbf{Metrics of the results:} In essence, face forgery detection is a two-class task, so we choose accuracy rate (ACC) and area under the receiver operating characteristic curve (AUC) to evaluate the performance of models. In the next tables, we calculate the frame and video level results in the cross-dataset evaluation. For video level, the same as other works \cite{Luo_2021_CVPR,masi2020two,F3-Net}, we average the prediction scores of all frames for each video as the final prediction of this video.

\begin{table*}[t]
	
	\begin{minipage}[t]{.5\textwidth}
		\begin{center}
			\begin{tabular}{c|cccc}
				\toprule
				Datasets & \multicolumn{4}{c}{Celeb-DF V2 \cite{celebdf}} \\
				\hline
				\multirow{2}{*}{Methods}
				& \multicolumn{2}{c}{Frame Level} & \multicolumn{2}{c}{Video Level} \\
				& ACC & AUC & ACC & AUC \\
				\hline
				\hline
				DDPGF \cite{sun2021improving}
				& - & 56.90  & - & - \\
				Two-branch \cite{masi2020two}
				& - & 73.41  & - & 76.65 -\\
				MADFD \cite{MADFD}
				& - & 67.44 & - & - \\
				DGFFD \cite{sun2021domain}
				& 63.40 & 64.10 & - & - \\
				F3-Net \cite{F3-Net}
				& - & 65.17 & - & - \\
				SPSL\_c23 \cite{SPSL}
				& - & 72.39 & - & - \\
				GFFD \cite{Luo_2021_CVPR}
				& - & - & - & 79.40 \\
				Xception
				& 69.65 & 67.58 & 72.66 & 73.54 \\
				ours
				& \textbf{72.69} & \textbf{74.55}  & \textbf{76.37} & \textbf{80.65} \\
				\bottomrule
			\end{tabular}
		\end{center}
		\caption{Cross-dataset evaluation on Celeb-DF V2 \cite{celebdf} dataset, whose model is trained on FF++ dataset. The ACC values of some methods are missing, so we mostly compare the performance with AUC.}
		\label{table:cross_cdf}
	\end{minipage}
	\hfill
	\begin{minipage}[t]{.45\textwidth}
		\begin{center}
			\begin{tabular}{c|cc|cc}
				\toprule
				Datasets       & \multicolumn{2}{c|}{FFIW}   & \multicolumn{2}{c}{WDF}   \\
				\hline
				Methods     & ACC & AUC   &  ACC & AUC   \\
				\hline
				\hline
				MesoNet \cite{afchar2018mesonet}
				&  53.80 & 55.40  & 64.47 & -  \\
				TSN \cite{wang2016temporal}
				&  61.10 & 62.80  & - & -  \\
				C3D \cite{tran2015learning}
				&  64.30 & 65.50  & 55.87 & -  \\
				I3D \cite{carreira2017quo}
				&  68.80 & 69.50  & 62.69 & -  \\
				FFIW-M \cite{zhou2021face}
				&  71.30 & 73.50  & -  & -  \\
				ADDNet \cite{zi2020wilddeepfake}
				&  -     & -     & 76.25 & -  \\
				Xception
				&  95.05 & 99.32  & 81.74 & 87.46  \\
				ours
				& \textbf{95.70} & \textbf{99.48} &  \textbf{82.72} & \textbf{89.96}  \\
				\bottomrule
			\end{tabular}
		\end{center}
		\caption{The results of within-dataset evaluation on FFIW and WDF datasets (video level). In this table, FFIW-M \cite{zhou2021face} and ADDNet \cite{zi2020wilddeepfake} mean the forgery detection methods proposed in FFIW \cite{zhou2021face} and WDF \cite{zi2020wilddeepfake}, respectively.}
		\label{table:self_ffiw_wdf}
	\end{minipage}
\end{table*}

\textbf{Within-dataset evaluation.} The results of within-dataset evaluation are presented in Table \ref{table:self_split}, which includes different compression ratio (c23 (high quality) and c40 (low quality)) and four manipulation methods of FF++. We can see that our method achieves the best or comparable performance under different settings. It is notable that on the harder datasets, e.g. FF++ with low quality, especially FS and NT, we achieve more significant improvement which verifies the learned feature by our method is more effective for different forgery methods and robust with quality variation. Compared with GFFD using SRM to obtain high frequency features, our method outperforms it by 6.6\% on FS, which shows multi-scale DWT is more effective for extracting the abundant high frequency information than these low-level filters. Compared with F3Net, SPSL, and FADFL adopting DFT or DCT to get frequency features, our method outperforms it by 5.3\%, 15.8\%, and 6.2\% on FF++\_c40, which verifies multi-scale DWT is more suitable for face forgery detection task due to its rich multi-scale high-frequency information. MADFD \cite{MADFD} utilized multiple attention maps and texture feature enhancement to capture local discriminative features. We utilize RGB and high frequency branches and fuse these features by FSF module. Compared with Two-branch and MADFD, it demonstrates that the proposed framework and FSF module are effective in capturing the forgery information. We also make a within-dataset evaluation on realistic scenarios i.e. FFIW and WDF datasets, whose results are shown in Table \ref{table:self_ffiw_wdf}. Our results are the best. So, the proposed method is more suitable for diverse deepfakes and real-world face forgery detection.

\textbf{Cross-dataset evaluation.} Celeb-DF V2 dataset is often used as cross-domain evaluation, so we evaluate the model on Celeb-DF V2, whose results are shown in Table \ref{table:cross_cdf}. Considering the diversity of realistic scenarios, we also make a cross-domain evaluation on WDF and FFIW datasets, whose results are shown in Table \ref{table:cross_ffiw}. From the Table \ref{table:cross_cdf} and \ref{table:cross_ffiw}, we can observe that our method achieves the state-of-the-art performance on Celeb-DF V2, FFIW, and WDF datasets. Therefore, MSWT is robust and effective in within- and cross-dataset, which demonstrates that multi-scale structure and FSF module can learn more forgery details and make full use of high-frequency and spatial features.

\begin{table}[htbp]
	\centering
	\begin{tabular}{c|cccc|cccc}
		\toprule
		Datasets & \multicolumn{4}{c|}{FFIW \cite{zhou2021face}} & \multicolumn{4}{c}{WDF \cite{zi2020wilddeepfake}} \\
		\hline
		\multirow{2}{*}{Methods}
		& \multicolumn{2}{c}{Frame Level} & \multicolumn{2}{c|}{Video Level} & \multicolumn{2}{c}{Frame Level} & \multicolumn{2}{c}{Video Level} \\
		& ACC & AUC & ACC & AUC & ACC & AUC & ACC & AUC \\
		\hline
		\hline
		Xception
		& 71.07 & 76.86 & 70.28 & 76.40 & 67.33 & 67.21 & 62.03 & 64.76\\
		ours
		& \textbf{73.11} & \textbf{81.54} & \textbf{76.10} & \textbf{82.68} & \textbf{68.55} & \textbf{68.71} & \textbf{63.28} & \textbf{67.30} \\
		\bottomrule
	\end{tabular}
	
	\caption{Cross-dataset evaluation on WildDeepfake (WDF) \cite{zi2020wilddeepfake} and FFIW \cite{zhou2021face} datasets, whose model is trained on FF++ dataset.}
	\label{table:cross_ffiw}
\end{table}

\subsection{Ablation Study and Analysis}

To demonstrate the effectiveness of the proposed framework, we do ablation studies both on frequency and spatial feature fusion (FSF) module and multi-scale high frequency representations. In the ablation study experiments, we use FF++ as the training dataset and test on FF++ and CelebDF datasets as within-dataset and cross-dataset evaluations.

\begin{figure*}[htbp]
	\centering
	\begin{minipage}[t]{.5\textwidth}
			
		\begin{tabular}{c|cc|cc}
			\toprule
			Test type & \multicolumn{2}{c|}{Self Eval} & \multicolumn{2}{c}{Cross Eval}  \\
			\hline
			\multirow{2}{*}{Methods} & \multicolumn{2}{c|}{FF++} & \multicolumn{2}{c}{Celeb-DF V2}  \\
			& ACC & AUC & ACC & AUC \\
			\hline
			\hline
			Xception        & 95.73 & 96.30 & 69.65 & 67.58 \\
			\hline
			Xception+DWT    & 96.85 & 99.35 & 71.45 & 69.22 \\
			\hline
			Xception+SA     & 96.29 & 99.19 & 72.54 & 71.88 \\
			\hline
			Xception+FSA    & 97.02 & 99.47 & 72.98 & 73.72 \\
			\hline
			Xception+CMA    & 97.09 & 99.41 & 72.64 & 73.70 \\
			\hline
			MSWT     & \textbf{97.23} & \textbf{99.48}  & \textbf{72.69} & \textbf{74.55} \\
			\bottomrule
		\end{tabular}
		\captionof{table}{The results of ablation study on frequency and spatial feature fusion (FSF) module. The metrics are ACC and AUC (frame-level).}
		\label{table:abla_freq}
	\end{minipage}
	\hfill
     \begin{minipage}[t]{.45\textwidth}
		
		\includegraphics[width=0.7\columnwidth]{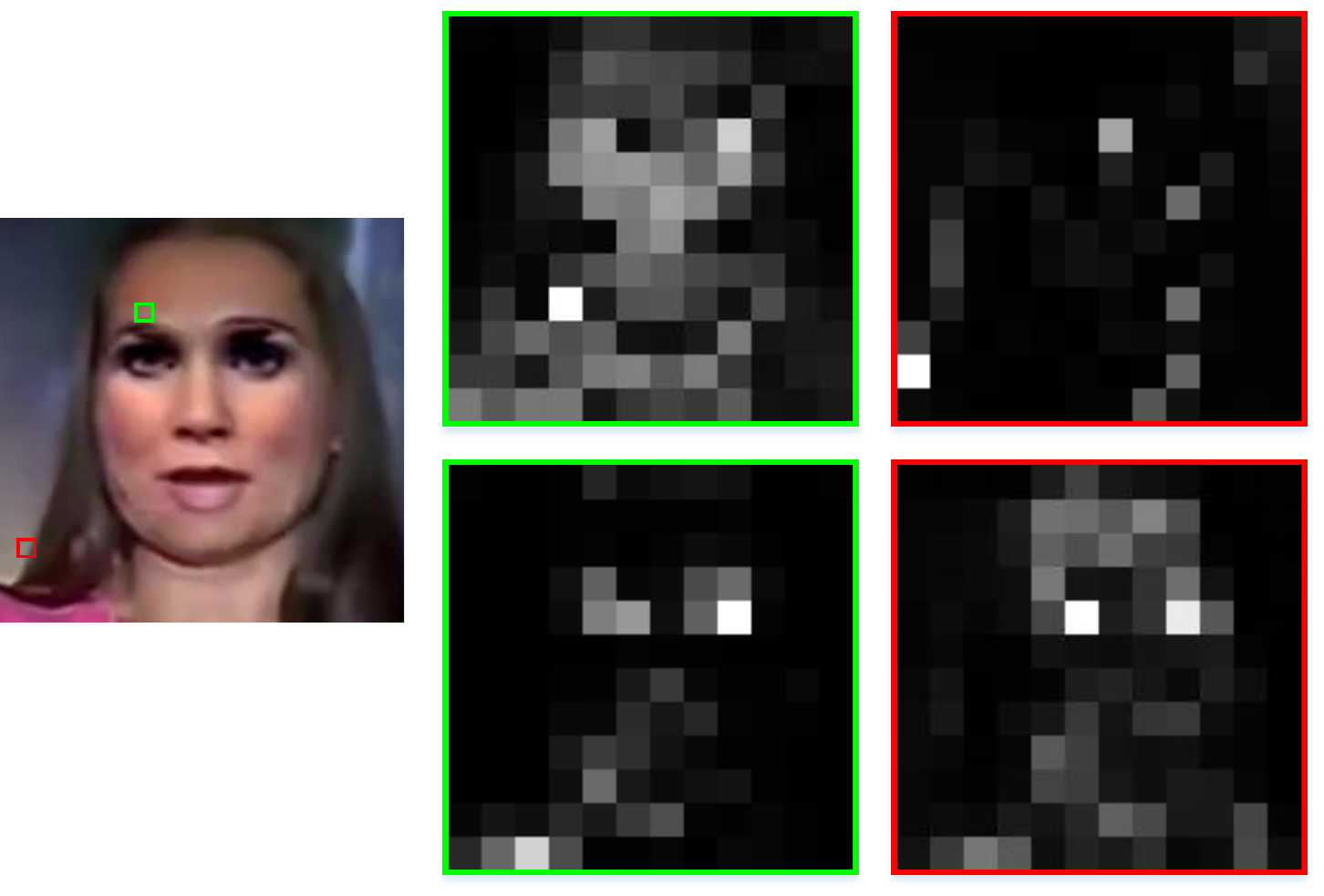}
		\caption{Visualization of multi-head attention map in FSA and SA shown in the 1st and 2nd rows. We choose two patches (a manipulated patch shown in green box) and a real patch shown in red box) to illustrate the difference between attention maps calculated by FAS and SA respectively.}
		\label{fig:att_patch_show}
	\end{minipage}
\end{figure*}

\textbf{Ablation study on frequency fusion.} The frequency and spatial feature fusion (FSF) module consists of a frequency-based spatial attention (FSA) and a cross-modality attention (CMA). Therefore, we keep single attention (FSA or CMA) to train the framework. To demonstrate the effectiveness of high-frequency features in FSA module, we utilize only RGB features to calculate attention, which is denoted as Xception+SA, and Xception+FSA means calculating the attention by frequency features. Besides, we train the Xception backbone by combining the high frequency sub bands into the corresponding stage directly without attention,named Xception+DWT in Table \ref{table:abla_freq}.

\begin{table}[htbp]
	\centering
	\begin{tabular}{c|cc|cc}
		\toprule
		Test type & \multicolumn{2}{c|}{Self Eval} & \multicolumn{2}{c}{Cross Eval}   \\
		\hline
		\multirow{2}{*}{DWT\_level} & \multicolumn{2}{c|}{FF++} & \multicolumn{2}{c}{Celeb-DF} \\
		& ACC & AUC & ACC & AUC \\
		\hline
		\hline
		Level 1       & 97.07 & 99.40     & 72.14 & 72.17 \\
		\hline
		Level 2       & 97.14 & 99.39     & 72.50 & 73.10 \\
		\hline
		Level 3 (MSWT) & \textbf{97.23} & \textbf{99.48} & \textbf{72.69} & \textbf{74.55} \\
		\bottomrule
	\end{tabular}
	\caption{The results of ablation study on multi-level structure of DWT. The metrics are ACC and AUC (frame-level).}
	\label{table:abla_level}
\end{table}
The results are shown in Table \ref{table:abla_freq}. We can observe that the result of Xception+DWT is better than Xception's, which means that the high frequency is useful for face forgery detection. Except for Xception, Xception+DWT which combines the high-frequency information directly has the worst results due to the misalignment between the frequency and RGB spaces. The results of Xception+SA are worse than Xception+FSA, which demonstrates that it is more effective to use the high frequency information to guide the attention calculation. The results of the method utilizing FSA or CMA are better than Xception+DWT, which demonstrates the effectiveness of the frequency-based spatial attention and cross-modality attention modules. The network with FSA and CMA achieves the best performance, which demonstrates that the two fusion modules are complementary.

To illustrate the influence of FSA and SA, we make a visualization of the attention in the 1st and 2nd second rows of Figure \ref{fig:att_patch_show}. We can see that the attention map of the manipulated region (the green rectangle) calculated via frequency features has high values in the manipulated region. The attention map of the real region (the red rectangle) is sparse in the manipulated region. While the attention maps calculated via spatial features have little difference between the manipulated and real regions, which learns less distinguish forgery details. Therefore, when using frequency features to guide the attention calculation, it effectively extracts the discriminable information between the real and fake regions.

\begin{figure}[htbp]
	\centering
	\begin{tabular}{c@{\hskip 1mm}c@{\hskip 1mm}c@{\hskip 1mm}c@{\hskip 1mm}c}
		\rotatebox{90}{~~~~Level 1}&
		\includegraphics[width=0.14\textwidth]{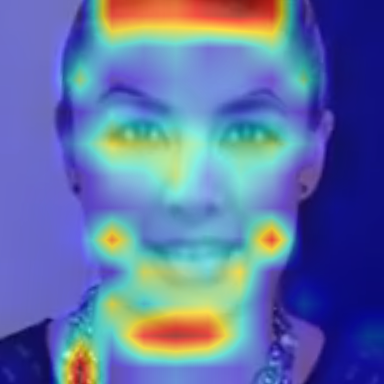}
		&
		\includegraphics[width=0.14\textwidth]{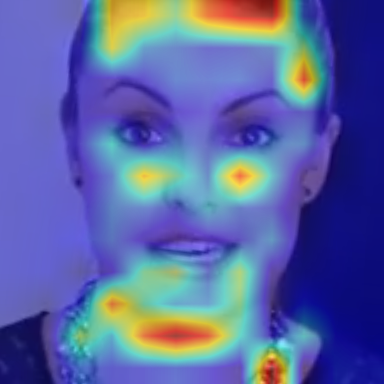}
		&
		\includegraphics[width=0.14\textwidth]{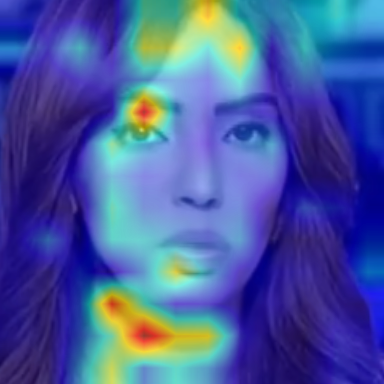}
		&
		\includegraphics[width=0.14\textwidth]{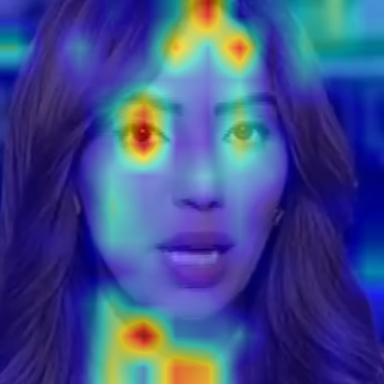}
		\\
		\rotatebox{90}{~~~~Level 2}&
		\includegraphics[width=0.14\textwidth]{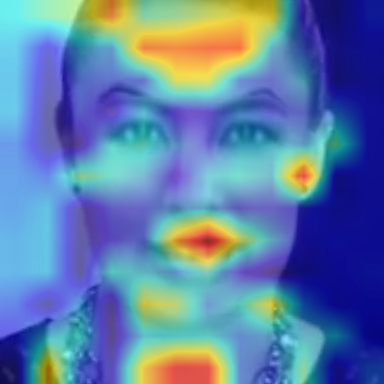}
		&
		\includegraphics[width=0.14\textwidth]{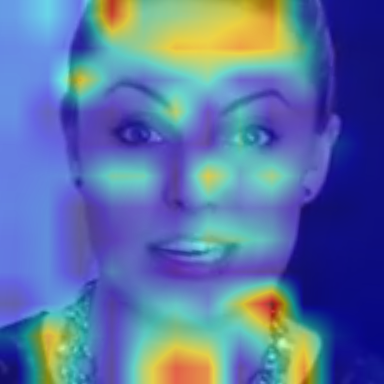}
		&
		\includegraphics[width=0.14\textwidth]{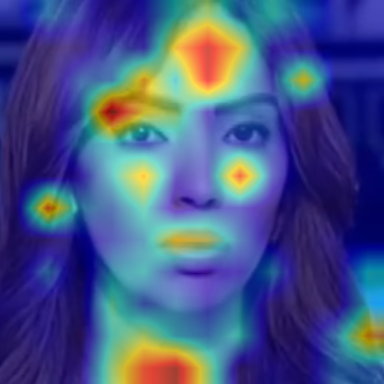}
		&
		\includegraphics[width=0.14\textwidth]{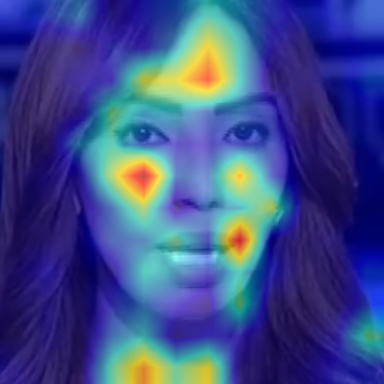}
		\\
		\rotatebox{90}{~~~~Level 3}&
		\includegraphics[width=0.14\textwidth]{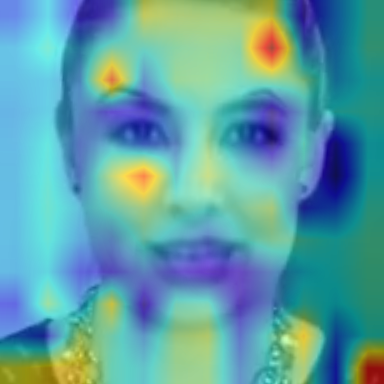}
		&
		\includegraphics[width=0.14\textwidth]{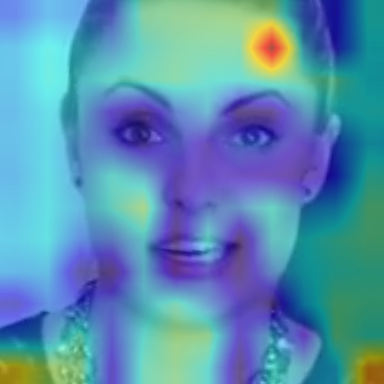}
		&
		\includegraphics[width=0.14\textwidth]{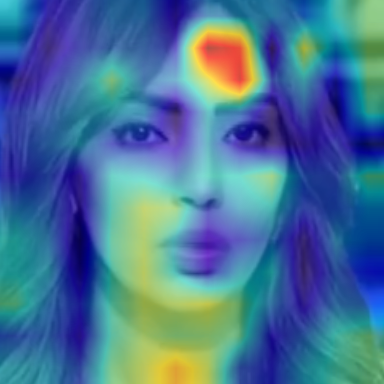}
		&
		\includegraphics[width=0.14\textwidth]{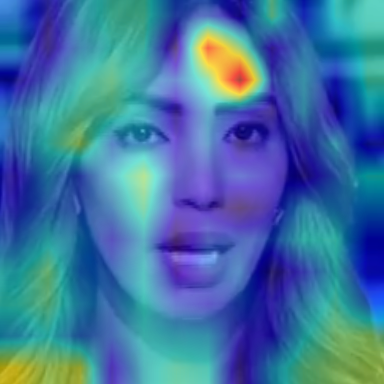}
		\\
		~ & fake & real & fake & real
	\end{tabular}
	\caption{Visualization of multi-level attention. The 1st and 2nd columns are fake images from DF and the corresponding real images. The third and fourth columns are fake images from NT and the corresponding real image. In this figure, the first to the third rows represent attention at level 1, 2, and 3 respectively.}
	\label{fig:att_vis}
\end{figure}

\textbf{Ablation study on multi-scale high frequency structure.} To illustrate the influence and effectiveness of the multi-scale high-frequency fusion, we do level-by-level experiments. We not only show the quantitative accuracy in Table \ref{table:abla_level}, but also make a visualization of the attention map at each level shown in Figure \ref{fig:att_vis}. In Table \ref{table:abla_level}, level 1, level 2, and level 3 represent the number of DWT levels used in the framework.

The results of the multi-scale wavelet transformer are shown in Table \ref{table:abla_level}. The performance of three-level frequency is the best compared with the results of levels 1 and 2, which demonstrates that multi-scale fusion is beneficial for face forgery detection. So the gradual aggregating fusion of multi-scale wavelet representation at different stages of the network can take full advantage of the frequency information.

We make a visualization on each level in Figure \ref{fig:att_vis}, we use the method proposed in \cite{Chefer_2021_CVPR,Chefer_2021_ICCV} to generate the visualization maps. We choose two examples from ID replacement Deepfakes and expression modification NeuralTextures. The fake image from Deepfakes and the corresponding real image, the fake face from NeuralTextures and the corresponding real image are in the 1st to 4th columns of Figure \ref{fig:att_vis} respectively. The visualization maps show that the fusion module can learn more global information at the higher level, and at the first and second levels, the fusion module learns more details about the local region. So via multi-scale frequency representations and fusion, we can enhance the feature learning from both the global structure and the local details simultaneously, which is important to face forgery detection task.

\section{Conclusion}

Considering the multi-scale analysis property of DWT and making full use of frequency information, we extract the multi-level frequency representations via DWT and use these high frequency components to the proposed multi-scale wavelet transformer architecture for face forgery detection. We apply transformer as the attention
block to integrate the high-frequency and RGB spatial features at multiple levels. Specifically,  the frequency based spatial attention guides  spatial features to focus on forgery regions. The cross-modality attention is used to better fuse the frequency features with the RGB spatial features. The various experiments demonstrate that the proposed framework is effective and robust on self and cross datasets compared with the existing methods.



\bibliographystyle{splncs}
\bibliography{egbib}

\end{document}